\documentclass[conference]{IEEEtran}
\usepackage{blindtext, graphicx}

\ifCLASSINFOpdf

\else

\fi

\usepackage[cmex10]{amsmath}

\usepackage{multirow}

\usepackage[final]{pdfpages}

\usepackage[]{algorithm}
\usepackage[noend]{algpseudocode}

\usepackage{algorithm}
\usepackage[noend]{algpseudocode}

\makeatletter
\def\BState{\State\hskip-\ALG@thistlm}
\makeatother
\usepackage{tabto}

\usepackage{amssymb}
\usepackage{fancyhdr}
\usepackage{colortbl}

\let\emptyset\varnothing

\IEEEoverridecommandlockouts
\begin{document}

\title{UDE-III: An Enhanced Unified Differential Evolution Algorithm for Constrained Optimization Problems}
\author{\IEEEauthorblockN{Anupam Trivedi and Dikshit Chauhan}
\thanks{Anupam Trivedi is with the Department of Electrical and Computer Engineering of the National University of Singapore, Singapore, \textit{email:} eleatr@nus.edu.sg.}
\thanks{Dikshit Chauhan is with the Department of Mathematics $\&$ Computing, Dr. B.R. Ambedkar National Institute of Technology Jalandhar, India, \textit{e-mail:} dikshitchauhan608@gmail.com}
\thanks{This research is supported by the Energy Market Authority (EMA) of Singapore under its EDGE project (Award number: EDGE2-GC2022-008)}}
\maketitle

\begin{abstract}
In this paper, an enhanced unified differential evolution algorithm, named UDE-III, is presented for real parameter-constrained optimization problems. The proposed UDE-III is a significantly enhanced version of the Improved UDE (i.e., IUDE or UDE-II), which secured the 1st rank in the CEC 2018 competition on constrained real parameter optimization. UDE-II was inspired by the following state-of-the-art DE variants - CoDE, JADE, SaDE, DE with ranking-based mutation operator, SHADE, and C$^2$oDE. To design UDE-III, we extensively analyzed and targeted the weaknesses of UDE-II. Specifically, UDE-III uses three trial vector generation strategies - DE/rand/1, DE/current-to-rand/1, and DE/current-to-pbest/1. It is based on a dual population approach, and for each generation, it divides the current population into two sub-populations. The top sub-population employs all three trial vector generation strategies on each target vector, just like in CoDE. On the other hand, in the bottom sub-population, one trial vector generation strategy is implemented on each target vector. The bottom sub-population employs strategy adaptation, wherein the probability of the three trial vector generation strategies is adapted in every generation by learning from their experiences in generating promising solutions in the previous learning period. The mutation operation in UDE-III is based on ranking-based mutation. Further, it employs the parameter adaptation principle of SHADE. The constraint handling principle in UDE-III is based on a combination of the feasibility rule and epsilon-constraint handling technique as proposed in C$^2$oDE. We observed that stagnation is a major weakness of UDE-II. To overcome this weakness, we took inspiration from the best-discarded vector selection (BDVS) strategy proposed in the literature and integrated a novel strategy in UDE-III to address stagnation. Additionally, UDE-III considers the size of two sub-populations to be a design element, whereas in UDE-II, the two sub-populations were assumed to be of equal size. Moreover, in comparison to UDE-II, UDE-III improves upon the strategy adaptation, ranking-based mutation, and epsilon constraint handling component of the constraint handling technique. The proposed UDE-III algorithm is tested on the 28 benchmark 30D problems provided for the CEC 2024 competition on constrained real parameter optimization. UDE-III is compared with its predecessor, UDE-II, and the experimental results demonstrate the efficacy of the proposed algorithm in solving constrained real parameter optimization problems. 
\end{abstract}

\begin{IEEEkeywords}
Constrained optimization, differential evolution, parameter adaptation, stagnation, strategy adaptation.
\end{IEEEkeywords}


\section{Introduction}


Differential evolution (DE) is a simple yet powerful evolutionary algorithm EA proposed by Price and Storn \cite{cit:24} in 1995 for real parameter optimization. DE is different from traditional EAs in the sense that it perturbs the current generation population members with the scaled differences of randomly selected and distinct population members. Over the last two decades, DE has been applied to a plethora of real-world problems and classical benchmark problems, owing to its simplicity and robust nature. Interested readers are referred to \cite{cit:23, Das2016} for a comprehensive review of DE.

In this paper, we present a DE algorithm named Unified Differential Evolution III (UDE-III) to solve constrained real-parameter optimization problems. The roots of UDE-III lie in UDE \cite{UDE}, which was the runner-up algorithm in the 2017 constrained real parameter optimization competition, and Improved UDE (IUDE or UDE-II), which was the winner of the 2018 constrained real parameter optimization competition \cite{UDE-II}. UDE, UDE-II, and UDE-III share the following principles. They are all built on the philosophy of unified algorithm design, comprehensive offline + online learning, and black box thinking. 

Unified algorithm design implies that to design a powerful algorithm, it is important to pay attention to all the design elements of the algorithm, not just one or few. Design elements are the components (i.e., operators) of an algorithm or its parameters on which its performance is critically dependent. In other words, unified algorithm design means looking at the algorithm as a whole instead of strengthening just one or few design elements. UDE-II, for example, took into consideration the following design elements - 1) DE trial vector generation strategy/strategy pool, 2) strategy adaptation, 3) control parameter adaptation, 4) parent selection strategy, and 5) constraint handling technique. In UDE-III, we went a few steps further and considered additional important design elements. One of the design elements is the stagnation strategy. In UDE-III, we integrated a novel stagnation strategy that utilizes the large number of unsuccessful trial vectors, which are generally discarded in DE. Further, UDE-II is based on a dual population approach; however, it considers the size of the dual populations to be equal. In UDE-III, we also considered the size of the dual populations as a design element. Overall, UDE-III holistically focuses on each and every design element. 

Offline + online learning consists of two parts - offline and online learning. In our design philosophy, offline learning indicates that it is important for the algorithm designer to learn offline, i.e., from other relevant high-performing algorithms proposed in the literature. On the other hand, online learning means that the algorithm learns or improves in an online manner from the data that is generated during the evolution. In UDE-II, offline learning was implemented in the sense that its design is inspired by many state-of-the-art algorithms along different design elements. UDE-II's design leverages many existing DE algorithms: JADE's DE trial vector generation strategy \cite{JADE}, CoDE's strategy pool \cite{CoDE}, SaDE's strategy adaptation \cite{SaDE}, SHADE's parameter adaptation \cite{SHADE}, rank-based DE's parent selection strategy \cite{rank}, and C$^2$oDE's constraint handling technique \cite{C$^2$oDE}. In UDE-III, we extended offline learning, and to design the stagnation strategy, we took inspiration from the best-discarded vector selection (BDVS) strategy \cite{BDVS}. 

Online learning refers to an algorithm's ability to continuously improve by using the data that is generated during the optimization process. UDE-II achieved this by incorporating online strategy adaptation and parameter adaptation mechanisms. In UDE-III, we extended online learning and leveraged the large number of unsuccessful trial vectors that were generated during the evolution. Specifically, We incorporated an external archive to store the unsuccessful trial vectors that are generated during the evolution. The archive size is always maintained to be less than equal to the population size by using the superiority of feasibility principle. In the stagnation strategy, when a stagnated individual meets the conditions to be replaced, an individual is randomly selected from the external archive to replace the stagnated individual. This way, a large number of unsuccessful trial vectors that are generally discarded during the evolution are utilized to improve the performance of the algorithm online.

The black box thinking term is adopted from the famous book by Matthew Syed titled ``Black Box Thinking: Marginal Gains and the Secrets of High Performance'' \cite{BB}. In this book, Matthew Syed argues that in order to succeed, it is of utmost importance to analyze the failures and address them. In our algorithm design philosophy, black box thinking implies that in order to improve the performance of an algorithm, it is essential to determine its weaknesses and exhaustively target them. Thus, UDE-III has been designed by comprehensively analyzing the weaknesses of UDE-II and addressing them. In UDE-II, weaknesses were identified in strategy adaptation, ranking-based mutation, constraint handling, and in terms of stagnation (getting stuck in local optima). In UDE-III, these weaknesses are systematically addressed.

Overall, the proposed UDE-III is a significantly enhanced version of the Improved UDE (i.e., IUDE or UDE-II). UDE-III uses three trial vector generation strategies - DE/rand/1, DE/current-to-rand/1, and DE/current-to-pbest/1. It is based on a dual population approach and at each generation, it divides the current population into two sub-populations. In the top sub-population, it employs all three trial vector generation strategies on each target vector, just like in CoDE. On the other hand, in the bottom sub-population, one trial vector generation strategy is implemented on each target vector. However, in contrast to UDE-II, wherein the size of the dual populations was assumed to be equal, in UDE-III, the size of the dual populations is considered to be a design parameter. The bottom sub-population in UDE-III employs strategy adaptation, wherein the probability of the three trial vector generation strategies is adapted in every generation by learning from their experiences in generating successful/unsuccessful trial vectors across the whole population in the previous learning period. The mutation operation in UDE-III is based on ranking-based mutation. Further, it employs the parameter adaptation principle of SHADE. The constraint handling principle in UDE-III is based on a combination of the feasibility rule and epsilon-constraint handling technique as proposed in C$^2$oDE. We observed that stagnation is a major weakness of UDE-II. To overcome this weakness, we took inspiration from the best-discarded vector selection (BDVS) strategy proposed in the literature and integrated a novel strategy in UDE-III to address stagnation. In contrast to UDE-II, UDE-III improves upon the strategy adaptation, ranking-based mutation, and epsilon constraint handling component of the constraint handling technique and integrates a novel strategy to overcome stagnation. The proposed UDE-III algorithm is tested on the 28 benchmark 30D problems provided for the CEC 2024 competition on constrained real parameter optimization \cite{Comp-CEC2024}. The experimental results demonstrate that UDE-III is significantly better than its predecessor, UDE-II, in solving constrained real parameter optimization problems.

The rest of the paper is organized as follows. The basics of DE are described in Section II. The weaknesses of UDE-II and the scope of potential improvements are described in Section III. Some DE variants relevant to addressing the weaknesses of UDE-II are described in Section III. In Section IV, the proposed UDE-III algorithm is presented. The parameter settings and the experimental results are summarized in Section V. Finally, the paper is concluded in Section VI.

\section{Differential Evolution}
\subsection{Differential Evolution}
Differential evolution consists of three basic steps: mutation, crossover, and selection. The basics of DE are presented below.
 
\vspace{-0.1in}
\subsection{Initialization}
The initial population is generated randomly. Suppose the $k$-th individual of the population at generation $G$ is denoted by $x_{k,G}$ (where  $x_{k,G} = [x_{1,k,G},x_{2,k,G},...,x_{D,k,G}]$, $D$ being the number of dimensions or decision variables).
The $j$-th decision variable of the $k$-th individual is randomly initialized for the initial population (at $G = 1$) as
\begin{equation}
x_{j,k,1}=x_{j,min}+rand_{k,j} [0,1].(x_{j,max}-x_{j,min})                    
\end{equation}
where $x_{j,min}$ and $x_{j,max}$ are the minimum and maximum bounds of the $j$-th decision variable, respectively and $rand_{k,j} [0,1]$ is an uniformly distributed random number lying between $0$ and $1$

\vspace{-0.1in} 
\subsection{Mutation}

At generation $G+1$, corresponding to $k$-th individual in the population, $x_{k,G}$ (called target vector in DE literature), DE creates a mutant individual $v_{k,G+1}$ (where $v_{k,G+1} = [v_{1,k,G+1},v_{2,k,G+1},...,v_{D,k,G+1}]$, $D$ being the number of decision variables) through mutation. There are several DE variants in the literature and they differ mainly in the way mutation operation is executed. DE/rand/1 is one of the most popular classical DE variants and the equation for mutation operation in DE/rand/1 is as follows:
\begin{equation}
\small
DE/rand/1:\;{v_{k,G+1}} = {x_{r_1^k,G}} + F( {{x_{r_2^k,G}} - {x_{r_3^k,G}}} )
\end{equation}
where $r_1^k$, $r_2^k$ and $r_3^k$ are mutually exclusive and randomly chosen indices from $[1, N_p]$ and are also different from the base index $k$ (where $N_p$ is the population size). The scaling factor $F$ is a control parameter and lies in the range $[0, 2]$. A smaller value of $F$ promotes exploitation while a larger value of $F$ promotes exploration

\subsection{Crossover}
After generating the mutant individual $v_{k,G+1}$ through mutation, crossover comes into operation. The two most popular crossover methods in DE literature are the binomial crossover and the exponential crossover. Here, we describe the binomial crossover. 
In binomial crossover, the mutant individual $v_{k,G+1}$ exchanges its components with the target individual $x_{k,G}$ with a probability $CR \in[0,1]$ to form the trial individual $u_{k,G+1}$ (where $u_{k,G+1} = [u_{1,k,G+1},u_{2,k,G+1},...,u_{D,k,G+1}]$, $D$ being the number of decision variables), according to the following condition:
\begin{equation}
\small
{u_{j,k,G+1}} = \left\{ {\begin{array}{*{20}{c}}
{{v_{j,k,G+1}}}&{if\left( {ran{d_{k,j}}[0,1] \le CR{\rm{\;or\;}}j = {j_{rand}}} \right)}\\
{{x_{j,k,G}}}&{otherwise}
\end{array}} \right.
\end{equation}
where $ran{d_{k,j}}[0,1]$ is a uniformly distributed random number and ${j_{rand}} \in [1,2,. . .,D]$ is a randomly chosen index which ensures that the trial individual gets at least one component from the mutant individual.

\subsection{Selection}
In the selection or the replacement step, a one to one comparison is performed between the target vector and the trial vector, and the fitter one survives into the next generation as shown below.
\begin{equation}
      x_{i,G+1}=
      \begin{cases}
u_{i,G+1},\ if\ f(u_{i,G+1})\leq f(x_{i,G})\\
x_{i,G},\ if\ f(u_{i,G+1})>f(x_{i,G})
      \end{cases}
\end{equation}

\section{Weaknesses of UDE-II and Scope of Potential Improvements}
In this section, the weaknesses of UDE-II are discussed in detail. 

\begin{itemize}
    \item \textbf{Size of the sub-populations} - In UDE-II, the size of the top sub-population and the bottom sub-population was assumed to be equal. However, during the design of UDE-III, we realized that the size of the sub-populations is itself a design element and it is incorrect to randomly assume equal size for the two sub-populations.
    
    \item \textbf{Ranking-based mutation} - Inspired by the performance of the ranking-based parent selection for mutation proposed by Gong and Cai, it was incorporated in UDE-II. However, in UDE-II, the original parent selection method proposed by Gong and Cai was modified in the sense that the base vector and the terminal vector in the mutation operation are selected from the top $NP/2$ members of the current population (i.e., from the top sub-population A, where $NP$ is the population size). We realized that selecting the parents from only the top sub-population worked well in UDE-II as the size of the two sub-populations was equal. However, if the size of the top sub-population is smaller that the size of the bottom sub-population, then ranking-based parent selection for mutation from only the top sub-population, may lead to over exploitation and can result in inferior performance of the algorithm.
    
    \item \textbf{Strategy Adaptation} - The strategy adaptation in UDE-II works as follows. In the top sub-population A, the three trial vectors generated corresponding to each target vector are compared among themselves, and the trial vector generation strategy with the best trial vector scores a win. At every generation, the success rate of each trial vector generation strategy is evaluated over the period of previous $Lp$ generations (where $Lp$ is the learning period). For example, the success rate ($SR$) of trial vector generation strategy 1 is given by $SR_1 = NW_1/(NW_1 + NW_2 + NW_3)$, where $NW_1$ is the number of wins of trial vector generation strategy over the learning period.
    
    In the bottom sub-population B, which consists of $NP/2$ members, corresponding to each target vector, a trial vector generation strategy is probabilistically employed. The probability of employing a trial vector generation strategy in the bottom sub-population B is equal to its recent success rate in the top sub-population A, i.e., the success rate evaluated over the previous learning period. 
    
    We reviewed the strategy adaptation in UDE-II, and realized it has two weaknesses. First of all, it considers the performance of the mutation strategies only in the top sub-population and ignores the valuable information about the performance of the strategies in the bottom sub-population. Secondly, the success rate of the mutation strategies is evaluated by only considering the number of wins in the learning period while the number of losses are completely ignored. This is unlike the strategy adaptation proposed in the popular SaDE algorithm from which UDE-II is inspired.
    
    \item \textbf{Data-driven Learning} - In UDE-II, like in other evolutionary algorithms, a lot of data is generated during the evolution. Though parameter adaptation and strategy adaptation uses lot of data to improve the algorithm's performance yet there are studies which have shown far more scope to exploit the useful data and extract learning. For example, Ghosh \textit{et al.} demonstrated that archiving the most promising difference vectors from past generations and then by reusing them for generating offspring in the subsequent generations, can strikingly improve the performance of DE.
    For example, a lot of trial vectors generated may be unsuccessful in every generation. Traditionally, these unsuccessful trial vectors are eliminated from the algorithm altogether. Thus, many function evaluations that are spent in generating these trial vectors go wasted. This is more critical in UDE-II as the top sub-population generates three trial vectors corresponding to every target vector. Hence, a lack of utilization of the large number of unsuccessful trial vectors though does not represent a weakness, it indicates an opportunity for potential improvement.
\end{itemize}

\section{Related Differential Evolution Variants}

The proposed improved unified differential evolution, termed UDE, is inspired by CoDE \cite{CoDE}, JADE \cite{JADE}, SaDE \cite{SaDE}, ranking-based mutation DE \cite{rank}, SHADE \cite{SHADE}, and C$^2$oDE \cite{C$^2$oDE}. The short summary of these algorithms is available in \cite{UDE-II}. Since, the stagnation strategy is the highlight of UDE-III, in this section, we discuss some related works which addressed the stagnation issue in DE.

Stagnation is a well-known problem in DE. In the evolution of DE algorithms, many individuals may converge to local optima. When individuals get stuck at local optima, many a times they stagnate, which means even after computational resources are allocated to them, they may not improve. This has considerable negative impact on the performance of DE. Thus, there have been several important research works on overcoming stagnation issue in DE.  
    
Zeng \textit{et al.} proposed a target vector replacement (TVRS) strategy to address the stagnation issue in DE \cite{TVRS}. The TVRS strategy is based on the idea that because improving stagnant individuals is difficult, it is better to provide more opportunities for improvement to non-stagnant individuals. TVRS framework consists of the following two operations which are performed alternately when it detects an individual to be in a stagnant state: 1) It randomly selects a non-stagnant individual from the population, and replaces the stagnant individual in order to carry out mutation operation and crossover operations; (2) It performs mutation and crossover operations with the stagnant individual. 

In \cite{BDVS}, Zeng \textit{et al.} presented a best discarded vector selection (BDVS) strategy to overcome the stagnation problem in DE. In BDVS, an external archive is created to record the best discarded trial vectors for each individual. During the replacement step, if a trial vector is worse than the target vector and the target vector is in a stagnant state, then the target vector is replaced with the best vector of the target vector or a vector chosen at random from the external archive. BDVS is a generic framework and can be easily integrated into other DE variants.

\section{Proposed Unified Differential Evolution-III}
In this section, the proposed algorithm UDE-III is presented. The proposed UDE-III is a significantly enhanced version of UDE-II (Improved UDE or IUDE) which secured the 1st rank in the CEC 2018 competition on constrained real parameter optimization \cite{UDE-II}. The components of UDE-III are described in detail below.



\subsection{Strategy Pool} 
The strategy pool in UDE-III consists of DE/rand/1, DE/current-to-rand/1, and DE/current-to-$p$best/1 trial vector generation strategies just like in UDE-II. 


\subsection{Crossover}
Binomial crossover and exponential crossover are two of the most popular crossover operators used in differential evolution variants. However, exponential crossover operator has been found to outperform binomial crossover operator in solving hard high dimensional problems for several large scale continuous optimization problems. Thus, UDE-III employs binomial crossover operator for problems with dimensions less than 100 while for problems with dimension 100 and above, it employs exponential crossover. However, as the CEC 2024 competition involves problems with 30D size, UDE-III uses binomial crossover.

\subsection{Dual Population}

In UDE-III, at each generation, the current population is divided into two sub-populations like in UDE-II. However, the two sub-populations are not randomly assumed to be of equal size like in UDE-II. In UDE-III, we considered the size of the two sub-populations also as a design element. At each generation, the current population is divided into two sub-populations (top sub-population A of size $T$ and bottom sub-population B of size $NP - T$. In the top sub-population A, each trial vector generation strategy is employed corresponding to each target vector, resulting in three trial vectors. However, for efficient allocation of computational resources, for the bottom sub-population B, strategy adaptation is utilized. 

\subsection{Strategy Adaptation}
In the top sub-population A, all three trial vector generation strategies are deployed for each target vector to generate three trial vectors like in CoDE. The three trial vectors generated corresponding to each target vector are compared among themselves, and the trial vector generation strategy with the best trial vector scores a win. The other two trial vector generation strategies are allocated losses. In the bottom sub-population B, corresponding to each target vector, a trial vector generation strategy is probabilistically employed. If the trial vector is better than the corresponding target vector, it scores a win otherwise it scores a loss. The probability of employing a trial vector generation strategy is equal (i.e., 0.33) for the first $Lp$ generations. Thereafter, in every generation, the probability of employing a trial vector generation strategy is calculated using the same method as proposed in SaDE. The probability evaluation method considers the number of wins and losses of each trial vector generation strategy over the whole population across the previous learning period $Lp.$



\begin{table*}[t]
\renewcommand{\arraystretch}{1}
  \centering
  \caption{Results obtained for 30$D$ ($MaxFES = 20000 \times D$)}
\resizebox{1\linewidth}{!}{\begin{tabular}{|l|c|c|c|c|c|c|c|}
    \hline
\bf	Func	&	\bf	F01	&	\bf	F02	&	\bf	F03	&	\bf	F04	&	\bf	F05	&	\bf	F06	&	\bf	F07	\\\hline
\bf	Best	&		0	&		0	&		0	&		0	&		0	&		0	&		-915.1	\\
\bf	Mean	&		1.57E-28	&		1.39E-28	&		92.208	&		6.5149	&		0.15946	&		0	&		-673.22	\\
\bf	Worst	&		4.86E-28	&		3.54E-28	&		236.21	&		13.573	&		3.9866	&		0	&		-225.63	\\
\bf	STD	&		9.90E-29	&		9.49E-29	&		56.503	&		6.9208	&		0.79732	&		0	&		151.67	\\
\bf	SR	&		100	&		100	&		100	&		100	&		100	&		100	&		100	\\
\bf	$\bar{Viol}$	&		0	&		0	&		0	&		0	&		0	&		0	&		0	\\
\bf	c	&		[000]	&		[000]	&		[000]	&		[000]	&		[000]	&		[000]	&		[000]	\\
\bf	$\bar{v}$	&		0	&		0	&		0	&		0	&		0	&		0	&		0	\\\hline
\bf	Func	&	\bf	F08	&	\bf	F09	&	\bf	F10	&	\bf	F11	&	\bf	F12	&	\bf	F13	&	\bf	F14	\\\hline
\bf	Best	&		-0.00028398	&		-0.00266550	&		-0.00010284	&		-0.83257	&		3.9825	&		0	&		1.4085	\\
\bf	Mean	&		-0.00028398	&		-0.00266550	&		-0.00010284	&		-197.96	&		3.9924	&		0.47839	&		1.4085	\\
\bf	Worst	&		-0.00028398	&		-0.00266550	&		-0.00010284	&		-1458.2	&		4.0639	&		3.9866	&		1.4085	\\
\bf	STD	&		4.92E-12	&		0	&		1.80E-14	&		499.63	&		0.022254	&		1.3222	&		4.35E-16	\\
\bf	SR	&		100	&		100	&		100	&		28	&		100	&		100	&		100	\\
\bf	$\bar{Viol}$	&		0	&		0	&		0	&		22.557	&		0	&		0	&		0	\\
\bf	c	&		[000]	&		[000]	&		[000]	&		[001]	&		[000]	&		[000]	&		[000]	\\
\bf	$\bar{v}$	&		0	&		0	&		0	&		9.70E-05	&		0	&		0	&		0	\\\hline
\bf	Func	&	\bf	F15	&	\bf	F16	&	\bf	F17	&	\bf	F18	&	\bf	F19	&	\bf	F20	&	\bf	F21	\\\hline
\bf	Best	&		2.3561	&		0	&		0.031047	&		36.52	&		0	&		1.4065	&		3.9825	\\
\bf	Mean	&		2.3561	&		0	&		0.68103	&		36.548	&		0	&		1.8511	&		9.2796	\\
\bf	Worst	&		2.3561	&		0	&		0.86878	&		37.217	&		0	&		2.4427	&		39.655	\\
\bf	STD	&		1.44E-06	&		0	&		0.44217	&		0.13941	&		0	&		0.28827	&		8.5072	\\
\bf	SR	&		100	&		100	&		0	&		100	&		0	&		100	&		100	\\
\bf	$\bar{Viol}$	&		0	&		0	&		14.708	&		0	&		21375	&		0	&		0	\\
\bf	c	&		[000]	&		[000]	&		[100]	&		[000]	&		[100]	&		[000]	&		[000]	\\
\bf	$\bar{v}$	&		0	&		0	&		14.5	&		0	&		21375	&		0	&		0	\\\hline
\bf	Func	&	\bf	F22	&	\bf	F23	&	\bf	F24	&	\bf	F25	&	\bf	F26	&	\bf	F27	&	\bf	F28	\\\hline
\bf	Best	&		7.98E-19	&		1.4085	&		2.3561	&		0	&		0.15774	&		36.52	&		65.457	\\
\bf	Mean	&		25.558	&		1.4502	&		2.3561	&		0.25132	&		0.68935	&		37.699	&		78.711	\\
\bf	Worst	&		165.69	&		1.4954	&		2.3561	&		6.283	&		1.0027	&		49.763	&		109.88	\\
\bf	STD	&		49.444	&		0.044322	&		1.08E-07	&		1.2566	&		0.26803	&		3.2411	&		15.935	\\
\bf	SR	&		100	&		100	&		100	&		100	&		0	&		100	&		0	\\
\bf	$\bar{Viol}$	&		0	&		0	&		0	&		0	&		15.277	&		0	&		21441	\\
\bf	c	&		[000]	&		[000]	&		[000]	&		[000]	&		[100]	&		[000]	&		[100]	\\
\bf	$\bar{v}$	&		0	&		0	&		0	&		0	&		15.5	&		0	&		21440	\\\hline
 \end{tabular}}
\end{table*}
\subsection{Constraint Handling} The constraint handling technique in UDE-III is similar to that in UDE-II. It is inspired by the combination of $\epsilon$ constraint handling technique \cite{eps} and superiority of feasible (SOF) solutions method \cite{SOF} proposed in C$^2$oDE \cite{C$^2$oDE}. The top sub-population in UDE-III employs exactly the same constraint handling technique as in C$^2$oDE \cite{C$^2$oDE} because the evolution of solutions in the top sub-population in UDE-III is similar to that in C$^2$oDE. However, as the bottom sub-population generates only one trial vector corresponding to each target vector, only the $\epsilon$ constraint handling technique \cite{eps} is employed in the bottom sub-population. 

The parameter settings for the $\epsilon$ constraint handling technique follow the same values as used in  C$^2$oDE \cite{C$^2$oDE} UDE-II except for the parameter $cp$. Our experimental investigations demonstrated that the value of $cp$ may vary considerably as $cp$ is derived from the following equation:

\begin{equation}
    cp = - (log \epsilon{_0} + \lambda)/(log(1-p)
\end{equation}
Here, $\epsilon_0$ is the initial threshold and set to be the maximum degree of constraint violation of the initial population. The maximum degree of constraint violation of the initial population i.e., $\epsilon_0$ may vary significantly depending upon the problem characteristics. As a result, $cp$ may also vary substantially. We observed that in some problems, $cp$ may take very high values because of which $\epsilon$ may decrease at a rapid exponential rate. This has a negative impact on the performance of UDE-III on several problems. Our experimental investigation demonstrated that limiting $cp$ to 33 significantly improved the performance of UDE-III. Hence, in UDE-III, we implement the condition that if $cp \ge$ 33 then $cp$ = 33.

\subsection{Parameter Adaptation}
The parameter adaptation in UDE-III is same as that in UDE-II. In particular, UDE-III employs SHADE \cite{SHADE} style parameter adaptation. Since the strategy pool of UDE-III consists of three strategies, it employs one pair of memory $M_F$ and $M_C$ for adaptation of $F$ and $CR$ corresponding to DE/rand/1/bin and DE/current-to-$p$best/1, and utilizes memory $M_F$ for adaptation of $F$ corresponding to DE/current-to-rand/1. However, the parameter adaptation proposed in SHADE \cite{SHADE} is suited only for bound-constrained i.e., unconstrained optimization. Thus, UDE-III employs the parameter adaptation as proposed in LSHADE44 \cite{LSHADE44}, which extends the SHADE \cite{SHADE} style parameter adaptation to constrained optimization problems. It is noted that only the top sub-population is utilized for the parameter adaptation in UDE-III just like in UDE-II. This is a weakness of UDE-III which could be addressed in the future to further improve the algorithm.  

\subsection{Migration between Sub-populations}
At the beginning of every generation, the population members are sorted according to the superiority of feasible solutions method. This step leads to migration of fitter (in terms of feasibility) solutions from bottom sub-population B to top sub-population A, and migration of less fit solutions from top sub-population A to bottom sub-population B. The advantage of migration is that the CoDE style trial vector generation strategy (i.e., implementation of three strategies) always gets implemented on the top sub-population members. Since, UDE-III employs DE/current-to-rand/1, and DE/current-to-$p$best/1 trial vector generation strategies, the CoDE style trial vector generation strategy being implemented on the best population members (in terms of feasibility) can promote exploitation of the fitter feasible solutions, and lead to faster convergence. It is noted that this step is just like in UDE-II. 

\subsection{Parent Selection}
Gong and Cai \cite{rank} proposed the ranking-based parent selection for mutation to solve unconstrained optimization problems. UDE-II conveniently adopted the method presented by Gong and Cai \cite{rank} as as it sorts the population members according to the superiority of feasible solutions method at the beginning of every generation. However, UDE-II modified the ranking-based parent selection method by selecting parents based on ranking from only the top sub-population. 

As discussed earlier, we realized that selecting the parents from only the top sub-population worked well in UDE-II as the size of the two sub-populations was equal. However, particularly if the size of the top sub-population is smaller that the size of the bottom sub-population, then ranking-based parent selection for mutation from only the top sub-population, may lead to over exploitation and can result in inferior performance of the algorithm. Hence, in UDE-III, the whole population is considered for selecting parents based on ranking-based mutation.

\subsection{Stagnation Strategy}
To design an effective strategy to overcome stagnation, there are several important research questions that need to be addressed. First, when can an individual be considered to be stagnated. Generally, if an individual has not been replaced by its trial vector for many generations, it can be said to be stagnated. Thus, we used a parameter $SG$, which indicates the number of generations for which, if an individual has not shown improvement, it can be said to be stagnated. 

Second, should a stagnated individual be immediately replaced, or should the stagnation strategy be applied if a certain proportion of the population stagnates? We researched this extensively and observed that replacing a stagnated individual immediately does not result in consistent superior algorithm performance. Hence, we used a parameter $SProp$ to indicate the threshold for the proportion of the stagnated population beyond which the stagnation strategy is activated. For example, $SProp$ = 50\% activates the stagnation strategy only when the number of stagnated individuals in the population exceeds 50\% of population size $NP$. 

Third, the choice of an individual which replaces a stagnated individual is important. In UDE-III, we maintain an external archive of size $NP$ to store the unsuccessful trial vectors generated during the evolution. The external archive is updated at every generation and the size of archive is always maintained to be less than equal to $NP$ using the superiority of feasibility principle. Once the proportion of the stagnated population is greater than $SProp$ of $NP$, one stagnated individual is replaced at every generation with a randomly selected vector from the external archive. This way UDE-III effectively utilizes the large number of unsuccessful trial vectors in the stagnation strategy.

\subsection{Replacement} 
UDE-III employs the traditional one-to-one comparison between target vector and trial vector to select a solution for the next generation. It is noted that as discussed above in the constraint handling strategy, the one-to-one comparison and selection between the target vector and trial vector is based on the $\epsilon$ constraint handling method. 

\subsection{Memory}
Since UDE-III employs a combination of $\epsilon$ constraint handling technique \cite{eps} and superiority of feasible (SOF) solutions method \cite{SOF}, there is a possibility that the best solution in terms of SOF may get lost from the evolving population. Hence, UDE-III utilizes an external memory and at the end of every generation, the memory is updated with the best solution with respect to SOF.

\section{Modifications made in UDE-II to Design UDE-III}
In this section, we summarize the modifications made in UDE-II to deisgn UDE-III.
\begin{itemize}
    \item \textbf{Stagnation strategy} -  The introduction of a novel strategy to address the stagnation issue is one of the significant highlights of UDE-III.

    \item \textbf{Size of the sub-populations} - While designing UDE-III, we challenged the assumption that the two sub-populations should be of equal size. We extensively experimented with the size of the sub-populations, and observed that it is an important design element of UDE-III. Our experiments indicated that the top sub-population A and the bottom sub-population B of size $0.25*NP$ and $0.75*NP$, respectively works the best, with $NP$ being 100 for 30D problems in the CEC 2024 competition.

    \item \textbf{Ranking-based mutation} - In UDE-II, the ranking-based selection of the base vector and the terminal vector in the mutation operation is carried out from the top sub-population A. However, we observed that it leads to over exploitation of the top sub-population members and inferior performance of the algorithm, particularly when the top sub-population is of the size $0.25*NP$ as in UDE-III. Hence, in UDE-III, the ranking-based selection of the base vector and the terminal vector in the mutation operation is carried out from the whole population instead of just top sub-population.

    \item \textbf{Strategy adaptation} - To overcome the shortcomings of the strategy adaptation design element in UDE-II, we took the following two measures. First, for determining the adaptive probability of each mutation strategy in the bottom sub-population, we recorded and utilized not only the number of wins but also the number of losses of each mutation strategy in the learning period. Secondly, this information was recorded over the entire population instead of just top sub-population like in UDE-II. Thus, the strategy adaptation design element has been made more comprehensive in UDE-III.

    \item \textbf{Constraint handling} - The constraint handling technique in UDE-III is the same as in UDE-II except that in UDE-III, for the epsilon constraint handling, we implement the condition that if $cp \ge$ 33 then $cp$ = 33.

\end{itemize}
\begin{table}
 \caption{Comparison of UDE-III with UDE-II at 30$D$.}
    \label{tab: UDE-II 30 D}
\centering
    \begin{tabular}{|l|c|c|}\hline
\bf	Func	&			UDE-III			&			UDE-II			\\\hline
		&	\bf	Mean	$\pm$	\bf	STD	&	\bf	Mean	$\pm$	\bf	STD	\\\hline
\bf	F1	&		1.570E-28	$\pm$		9.900E-29	&	\cellcolor{gray}	4.130E-29	$\pm$		2.260E-29	\\\hline
\bf	F2	&		1.390E-28	$\pm$		9.490E-29	&	\cellcolor{gray}	4.420E-29	$\pm$		2.580E-29	\\\hline
\bf	F3	&	\cellcolor{gray}	9.221E+01	$\pm$		5.650E+01	&		1.291E+02	$\pm$		2.954E+01	\\\hline
\bf	F4	&	\cellcolor{gray}	6.515E+00	$\pm$		6.921E+00	&		1.367E+01	$\pm$		4.693E-01	\\\hline
\bf	F5	&		1.595E-01	$\pm$		7.973E-01	&	\cellcolor{gray}	5.710E-29	$\pm$		9.910E-29	\\\hline
\bf	F6	&	\cellcolor{gray}	0.000E+00	$\pm$		0.000E+00	&		$(12\%)$				\\\hline
\bf	F7	&	\cellcolor{gray}	-6.732E+02	$\pm$		1.517E+02	&		$(76\%)$				\\\hline
\bf	F8	&	\cellcolor{gray}	-2.840E-04	$\pm$		4.920E-12	&	\cellcolor{gray}	-2.840E-04	$\pm$		1.250E-12	\\\hline
\bf	F9	&	\cellcolor{gray}	-2.666E-03	$\pm$		0.000E+00	&	\cellcolor{gray}	-2.666E-03	$\pm$		0.000E+00	\\\hline
\bf	F10	&	\cellcolor{gray}	-1.028E-04	$\pm$		1.800E-14	&	\cellcolor{gray}	-1.028E-04	$\pm$		5.060E-15	\\\hline
\bf	F11	&		$(28\%)$				&	\cellcolor{gray}	$(92\%)$				\\\hline
\bf	F12	&		3.992E+00	$\pm$		2.225E-02	&	\cellcolor{gray}	3.983E+00	$\pm$		9.880E-05	\\\hline
\bf	F13	&	\cellcolor{gray}	4.784E-01	$\pm$		1.322E+00	&		3.542E+00	$\pm$		1.609E+01	\\\hline
\bf	F14	&	\cellcolor{gray}	1.409E+00	$\pm$		4.350E-16	&	\cellcolor{gray}	1.409E+00	$\pm$		1.050E-15	\\\hline
\bf	F15	&	\cellcolor{gray}	2.356E+00	$\pm$		1.440E-06	&		5.875E+00	$\pm$		3.550E+00	\\\hline
\bf	F16	&	\cellcolor{gray}	0.000E+00	$\pm$		0.000E+00	&		1.571E+00	$\pm$		5.120E-07	\\\hline
\bf	F17	&	\cellcolor{gray}	$(0\%)$	$\pm$		1.471E+01	&		$(0\%)$			1.534E+01	\\\hline
\bf	F18	&	\cellcolor{gray}	3.655E+01	$\pm$		1.394E-01	&		$(0\%)$			3.804E+03	\\\hline
\bf	F19	&	\cellcolor{gray}	$(0\%)$	$\pm$		2.138E+04	&	\cellcolor{gray}	$(0\%)$			2.138E+04	\\\hline
\bf	F20	&	\cellcolor{gray}	1.851E+00	$\pm$		2.883E-01	&		3.885E+00	$\pm$		2.835E-01	\\\hline
\bf	F21	&	\cellcolor{gray}	9.280E+00	$\pm$		8.507E+00	&		1.655E+01	$\pm$		1.088E+01	\\\hline
\bf	F22	&	\cellcolor{gray}	2.556E+01	$\pm$		4.944E+01	&		3.723E+01	$\pm$		5.697E+01	\\\hline
\bf	F23	&		1.450E+00	$\pm$		4.432E-02	&	\cellcolor{gray}	1.429E+00	$\pm$		3.789E-02	\\\hline
\bf	F24	&	\cellcolor{gray}	2.356E+00	$\pm$		1.080E-07	&		2.482E+00	$\pm$		6.283E-01	\\\hline
\bf	F25	&	\cellcolor{gray}	2.513E-01	$\pm$		1.257E+00	&		8.734E+00	$\pm$		4.948E+00	\\\hline
\bf	F26	&	\cellcolor{gray}	$(0\%)$	$\pm$		1.528E+01	&		$(0\%)$			1.550E+01	\\\hline
\bf	F27	&	\cellcolor{gray}	3.770E+01	$\pm$		3.241E+00	&		$(0\%)$			6.570E+03	\\\hline
\bf	F28	&	\cellcolor{gray}	$(0\%)$	$\pm$		2.144E+04	&		$(0\%)$			2.145E+04	\\\hline
 \end{tabular}
\end{table}
\section{Experimental Results}
The proposed UDE-III algorithm is tested on the 28 CEC 2024 benchmark-constrained optimization problems with 30 dimensions ($D$). The parameter settings of UDE-III are as follows:
\begin{itemize}
    \item Population size $NP$ = 100,
    \item Size of top sub-population $T$ = 25,
    \item Learning period $Lp$ = 25 generations,
    \item $SG$ = 35,
    \item $SProp$ = 50\%
\end{itemize}

On each problem, 25 independent runs of UDE-III are taken. The results obtained on 30$D$ issues are reported in Table I. The comparative results of UDE-III and UDE-II are reported in Table II. Table II shows that UDE-III is inferior, similar, and superior to UDE-II on 7, 5, and 16 functions, respectively. Overall, this depicts the effectiveness of UDE-III in comparison to UDE-II in solving constrained real-parameter optimization problems.

\section{Conclusion}
In this paper, an enhanced unified differential evolution algorithm, named UDE-III, has been proposed to solve real parameter constrained optimization problems of CEC 2024 competition. UDE-III is a significantly enhanced version of the  Improved UDE (i.e., IUDE or UDE-II) which secured the 1st rank in the CEC 2018 competition on constrained real parameter optimization. UDE-III is built on the philosophy of unified algorithm design, comprehensive offline + online learning, and black box thinking. UDE-III has been designed by targeting the weaknesses of UDE-II (i.e., by implementing black box thinking) and by focusing on each and every design element (i.e., by focusing on unified algorithm design). It leverages on many state-of-the-art algorithms proposed in the literature (i.e., offline learning). Moreover, it implements online learning in the form of strategy adaptation, parameter adaptation, and strategy to overcome stagnation. Specifically, in contrast to UDE-II, UDE-III improves upon the strategy adaptation, ranking-based mutation, epsilon constraint handling component of the constraint handling technique, and integrates a novel strategy to overcome stagnation. The comparative experimental results on the CEC 2024 test suite against UDE-II demonstrates the superiority of UDE-III in solving real parameter constrained optimization problems.


\bibliographystyle{IEEEtran}
\bibliography{references}

\begin{thebibliography}{10}
\providecommand{\url}[1]{#1}
\csname url@samestyle\endcsname
\providecommand{\newblock}{\relax}
\providecommand{\bibinfo}[2]{#2}
\providecommand{\BIBentrySTDinterwordspacing}{\spaceskip=0pt\relax}
\providecommand{\BIBentryALTinterwordstretchfactor}{4}
\providecommand{\BIBentryALTinterwordspacing}{\spaceskip=\fontdimen2\font plus
\BIBentryALTinterwordstretchfactor\fontdimen3\font minus \fontdimen4\font\relax}
\providecommand{\BIBforeignlanguage}[2]{{%
\expandafter\ifx\csname l@#1\endcsname\relax
\typeout{** WARNING: IEEEtran.bst: No hyphenation pattern has been}%
\typeout{** loaded for the language `#1'. Using the pattern for}%
\typeout{** the default language instead.}%
\else
\language=\csname l@#1\endcsname
\fi
#2}}
\providecommand{\BIBdecl}{\relax}
\BIBdecl

\bibitem{cit:24}
R.~Storn and K.~Price, ``Differential evolution - a simple and efficient adaptive scheme for global optimization over continuous spaces,'' International Computer Science Institute, Berkeley, CA, USA, 1995, TR-95--012, Tech. Rep., 1995.

\bibitem{cit:23}
S.~Das and P.~N. Suganthan, ``\BIBforeignlanguage{English}{Differential evolution: a survey of the state-of-the-art},'' \emph{\BIBforeignlanguage{English}{IEEE Transactions on Evolutionary Computation}}, vol.~15, no.~1, pp. 4 -- 31, 2011.

\bibitem{Das2016}
S.~Das, S.~S. Mullick, and P.~Suganthan, ``Recent advances in differential evolution – an updated survey,'' \emph{Swarm and Evolutionary Computation}, pp.~--, 2016.

\bibitem{UDE}
A.~Trivedi, K.~Sanyal, P.~Verma, and D.~Srinivasan, ``A unified differential evolution algorithm for constrained optimization problems,'' in \emph{2017 IEEE Congress on Evolutionary Computation (CEC)}, June 2017, pp. 1231--1238.

\bibitem{UDE-II}
A.~Trivedi, D.~Srinivasan, and N.~Biswas, ``An improved unified differential evolution algorithm for constrained optimization problems,'' IEEE Congress on Evolutionary Computation Conference 2018, Tech. Rep., 2018.

\bibitem{JADE}
J.~Zhang and A.~C. Sanderson, ``{JADE}: Adaptive differential evolution with optional external archive,'' \emph{IEEE Transactions on Evolutionary Computation}, vol.~13, no.~5, pp. 945--958, Oct 2009.

\bibitem{CoDE}
Y.~Wang, Z.~Cai, and Q.~Zhang, ``Differential evolution with composite trial vector generation strategies and control parameters,'' \emph{IEEE Transactions on Evolutionary Computation}, vol.~15, no.~1, pp. 55--66, Feb 2011.

\bibitem{SaDE}
A.~K. Qin, V.~L. Huang, and P.~N. Suganthan, ``Differential evolution algorithm with strategy adaptation for global numerical optimization,'' \emph{IEEE Transactions on Evolutionary Computation}, vol.~13, no.~2, pp. 398--417, April 2009.

\bibitem{SHADE}
R.~Tanabe and A.~Fukunaga, ``Evaluating the performance of {SHADE} on {CEC} 2013 benchmark problems,'' in \emph{2013 IEEE Congress on Evolutionary Computation}, June 2013, pp. 1952--1959.

\bibitem{rank}
W.~Gong and Z.~Cai, ``Differential evolution with ranking-based mutation operators,'' \emph{IEEE Transactions on Cybernetics}, vol.~43, no.~6, pp. 2066--2081, Dec 2013.

\bibitem{C$^2$oDE}
B.~C. Wang, H.~X. Li, J.~P. Li, and Y.~Wang, ``Composite differential evolution for constrained evolutionary optimization,'' \emph{IEEE Transactions on Systems, Man, and Cybernetics: Systems}, pp. 1--14, 2018.

\bibitem{BDVS}
Z.~Zeng, Z.~Hong, H.~Zhang, M.~Zhang, and C.~Chen, ``\BIBforeignlanguage{en}{Improving differential evolution using a best discarded vector selection strategy},'' \emph{\BIBforeignlanguage{en}{Inf. Sci.}}, vol. 609, pp. 353--375, Sep. 2022.

\bibitem{BB}
M.~Syed, \emph{Black box thinking: Marginal Gains and the Secrets of High Performance}.\hskip 1em plus 0.5em minus 0.4em\relax John Murray Publishers Ltd., London, 2016.

\bibitem{Comp-CEC2024}
Q.~Kangjia, W.~Xupeng, B.~Xuanxuan, C.~Peng, P.~Kenneth~V., S.~Ponnuthurai~N., L.~Jing, W.~Guohua, and Y.~Caitong, ``Evaluation criteria for {CEC} 2024 competition and special session on numerical optimization considering accuracy and speed,'' Zhengzhou University, Central South University, Henan Institute of Technology, Qatar University, Tech. Rep., 2018.

\bibitem{TVRS}
Z.~Zeng, M.~Zhang, Z.~Hong, H.~Zhang, and H.~Zhu, ``\BIBforeignlanguage{en}{Enhancing differential evolution with a target vector replacement strategy},'' \emph{\BIBforeignlanguage{en}{Comput. Stand. Interfaces}}, vol.~82, Aug. 2022.

\bibitem{eps}
T.~Takahama and S.~Sakai, ``Constrained optimization by the $\epsilon$ constrained differential evolution with an archive and gradient-based mutation,'' in \emph{IEEE Congress on Evolutionary Computation}, July 2010, pp. 1--9.

\bibitem{SOF}
K.~Deb, ``\BIBforeignlanguage{English}{An efficient constraint handling method for genetic algorithms},'' \emph{\BIBforeignlanguage{English}{Computer Methods in Applied Mechanics and Engineering}}, vol. 186, no. 2-4, pp. 311 -- 38, 2000.

\bibitem{LSHADE44}
R.~Poláková, ``L-{SHADE} with competing strategies applied to constrained optimization,'' in \emph{2017 IEEE Congress on Evolutionary Computation (CEC)}, June 2017, pp. 1683--1689.

\end{thebibliography}

\end{document}